\documentclass{article}

\usepackage{PRIMEarxiv}

\usepackage[utf8]{inputenc} 
\usepackage[T1]{fontenc}    
\usepackage{hyperref}       
\usepackage{url}            
\usepackage{booktabs}       
\usepackage{amsfonts}       
\usepackage{nicefrac}       
\usepackage{microtype}      
\usepackage{lipsum}
\usepackage{fancyhdr}       
\usepackage{graphicx}       
\graphicspath{{media/}}     

\usepackage[utf8]{inputenc}
\usepackage{amsmath}
\usepackage{algorithm}
\usepackage{algpseudocode}
\usepackage{booktabs}
\usepackage{graphicx}
\usepackage[strings]{underscore}
\DeclareMathSymbol{\mh}{\mathord}{operators}{`\-}
\usepackage[justification=centering]{caption}
    
\usepackage{PRIMEarxiv}

\usepackage[utf8]{inputenc} 
\usepackage[T1]{fontenc}    
\usepackage{hyperref}       
\usepackage{url}            
\usepackage{booktabs}       
\usepackage{amsfonts}       
\usepackage{nicefrac}       
\usepackage{microtype}      
\usepackage{lipsum}
\usepackage{fancyhdr}       
\usepackage{graphicx}       
\graphicspath{{media/}}     

\title{Development and Comparison of Scoring Functions in Curriculum Learning}

\author{
  H. Toprak Kesgin, M. Fatih Amasyali \\
  \textit{Department of Computer Engineering} \\
  \textit{Yildiz Technical University} \\
  Istanbul, Turkey\\
  \texttt{\{tkesgin, amasyali\}@yildiz.edu.tr}
}

\begin{document}
\maketitle

\begin{abstract}
Curriculum Learning is the presentation of samples to the machine learning model in a meaningful order instead of a random order. The main challenge of Curriculum Learning is determining how to rank these samples. The ranking of the samples is expressed by the scoring function. In this study, scoring functions were compared using data set features, using the model to be trained, and using another model and their ensemble versions. Experiments were performed for 4 images and 4 text datasets. No significant differences were found between scoring functions for text datasets, but significant improvements were obtained in scoring functions created using transfer learning compared to classical model training and other scoring functions for image datasets. It shows that different new scoring functions are waiting to be found for text classification tasks.
\end{abstract}


\section{Introduction}
Curriculum is very often used in human and animal learning \cite{krueger2009flexible, skinner1958reinforcement}. Curriculum refers to the ordering of the samples of the subjects to be learned from easy to difficult. The reason for using this order is that it presumably provides better and faster learning. Deep learning is inspired by human learning. However, in its classical form, it uses all samples completely randomly throughout the training.

It is suggested to start training with easy samples in machine learning algorithms. And the effectiveness of curriculum learning (CL) has been demonstrated in a variety of investigations, including natural language processing (NLP) tasks, image classification and artificial data sets. CL performed better on the test set. It also accelerated the convergence speed \cite{bengio2009curriculum}. 

In deep neural network training, parameter updates are performed using the entire data set at each epoch. In order to apply the curriculum, first of all, all samples should be ordered (scoring function) and it should be determined at what rate (pacing function) the samples will be presented to the model. For example, let the total training set consist of 3n samples. After the samples are sorted according to a certain criterion, with the easiest n of them, parameter updates are performed as the number of epochs is determined. Then updates are performed for the easiest 2n samples and finally the training is completed using the entire dataset. The scoring function determines which samples will be used first, and the pacing function determines how fast the used subset size changes during training.

In order to apply CL in deep learning, a curriculum needs to be designed. It brings an additional computational load. Deep learning works better with large datasets. In addition to creating a dataset with a large number of samples, it will be very costly to determine the sample hardness of the dataset. For this reason, automatic curriculum methods for deep learning have been proposed.

For example, image classification experiments were carried out on artificially created geometric shapes \cite{bengio2009curriculum}. Shape sizes and object positions are used for sample difficulty. However, such methods are not applicable for all datasets. For this reason, methods that create a curriculum using a different machine learning model have been proposed. Predictions made by a trained model on a training set are used for curriculum. 

For natural language processing tasks, loss values on the training set of a different model can be used, as well as external methods. These methods include sentence length, word frequency in sentences, and n-gram based methods \cite{campos2021curriculum}. These features can be presented to the neural network in the order of smallest to largest, or vice versa. 

In this study, we compared automatic curriculum creation methods. We have shown which curriculum method works with experimental studies. For this, we used 4 image classification and 4 natural language processing datasets. We have also proposed a simple alternative to the existing self-thought method.

Ensemble learning is the simplest way to improve the success of machine learning models. That is, combining predictions from multiple models. We also used automatic curriculum methods in a hybrid structure in the study and examined the effect of the ensemble curriculum on success. In the experiments, we compared curriculum methods for all dataset-model pairs.

\section{Literature Review}

It has been demonstrated that using the CL approach for nonconvex optimization can result in significant improvements  \cite{bengio2009curriculum}. In the artificially created geometric shapes dataset experiments, the difficulty criterion was calculated based on the variances of the parameters such as object position and size of the created shapes. In language modelling trials, word frequency was determined as the difficulty determination criterion. Firstly, authors only used sentences with the most frequent 5000 words. Then added 5000 additional words and used the sentences that contain the first 10,000 words and so on until had used the entire data set. Authors demonstrated in both experiments that the success achieved with CL was statistically important when the entire data set was trained without the use of ordering.

For CL, the scoring function, which determines the difficulties of the samples, and the pacing function, which determines the speed at which the samples are presented to the model are defined \cite{hacohen2019power}. In experiments, authors used training set loss values according to the final parameters of the trained model. The ranking was determined using two separate models. The same model to be trained (self-thought) and a more complicated model. After one of the two models determines the scores, the model is trained from scratch using these scores. CL's performance is demonstrated in a variety of image classification datasets. CL obtained higher success in the test set and faster convergence. 

The training set's difficulty values do not have to be fixed before the training begins. SPL determines the difficulty values of the training set while training the model \cite{kumar2010self}. Difficulty values are recalculated at the end of each epoch based on the model's current hypothesis. Samples that are difficult at first may become easier over time, according to this method. SPL's success was experimentally demonstrated on four different data sets. 

The DIH approach is another method for dynamically determining the difficulty value of the training set \cite{zhou2020curriculum}. Unlike SPL, DIH uses not only the final version of the model but also its values in previous iterations, rather than only using the current hypothesis of the model. DIH was calculated by taking a moving average over the loss values created by the model throughout the training. Unlike other methods, the training set does not grow over time with this strategy. Training starts with the entire dataset and gradually decreases. Subsets are chosen  probabilistically based on their current DIH values. In 11 image classification datasets, DIHCL outperformed the classical random stochastic gradient descent (SGD) and recent CL methods. 

In NLP, difficulty determination can be made based on corpus characteristics, as well as using a different model. Sentence lengths can be used as difficulty values \cite{platanios2019competence}. It can be argued that long sentences are more difficult because they usually consist of short sentences. The second difficulty determination approach is Word Rarity. According to this method, the sentences in which the words rarely used in the corpus are stated as more difficult. The results obtained with these strategies drastically cut the training time. They were more successful than existing methods in NLU tasks.

Another method used to determine the difficulty values of sentences is The Sentence Entropy: N-gram difficulty method \cite{campos2021curriculum}.
 With the assumption, if a rare word is discovered early in the training, the learned word embeddings will have a high variance and will most likely have a poor representation. It is reported that although CL may provide a benefit to small datasets, this benefit is lost in large datasets.

\section{Method}

In this section, we compared the following strategies for obtaining scores in image classification datasets; using the model itself (self-thought), using a more complicated model (transfer scoring function), the test score version of these approaches, and lastly the ensemble versions of these methods. For NLP tasks, we investigated sentence length  and sentence entropy strategies, as well as the effects of combining these methods to create an ensemble curriculum. In addition, we compared the orders in our experiments by training them with two alternative curricula algorithms: Greedy-Curriculum Learning (GCL) and Probabilistic-Curriculum Learning (PCL). We compared 18 difficulty rankings-CL model pairings (including vanilla and random rankings) for the four image datasets and 17 difficulty rankings-CL model pairs for four text classifications in total. The remainder of this section describes these methods.

CL is based on the assumption that presenting simple samples to the model earlier will help in the training process. Using a meaningful ordering instead of randomly using all data can both shorten the training time and improve the classifier's final performance. 

Let $D = (x_i,y_i)$ where $D$ represents for training data, $x_i \in R^d$ single training point and $y_i$ its label. We defined the difficulty values of the training samples with the scoring function $f$. The main challenge for CL is to specify the f function. Scoring function can be any function, $(x_i,y_i)$ is more difficult than sample $(x_j,y_j)$ if f $(x_i,y_i) < f (x_j,y_j)$. Then we defined the pacing function, $g_t$. $g_t$ is used to determine the size of the training set's subsets $D'_1,...D'_n \subseteq D$ 

\subsection{Model Based Scoring Functions}

First, we used the model itself as a scoring method. In this method, the model is trained for the specified number of epochs. Then, the loss values are calculated for each training sample in the training set. These loss values are then inverted and converted to probability values. The model is then trained from the beginning with different initial weights using these scores. $k_i = \frac{1}{l(y_i,F_1(X_i,W)))}$ and $r_i = \frac{k_i}{\sum k_i} $ where $l$ is loss function, $ F_1 $ is trained model, $W$ is final model parameters for $ F_1 $ and $r$ is score values for each sample. If we used model itself to calculate scores, this method is called {\it self-thought } (ST) scoring function. Following the calculation of scores, the model is trained from scratch with different initial weights.

As a second method, we use a {\it transfer learning} (TL) scoring function to determine the scores. The transfer learning scoring function is calculated in the same way as self-thought, but a different model is used to obtain the scores. Usually, training to get the scores is done with a more successful and complicated model than $ F_1 $.  The $ F_1 $ model refers to the model to be trained.

Predictions generated by machine learning models on data that the model has not yet seen will provide more reliable results to the model than predictions made on data that the model has already been trained on. In both methods, the model is trained using the entire training set, and the predictions of the final hypothesis of the model are used. 

Therefore, we propose a method as a simple alternative to these two methods. Instead of training the model on the entire data set, the training data set is divided into two equal parts first. These two subsets are used to train two different models. The model predictions on the data set on which the model has not been trained are used to compute the difficulty scores. Scores are generated on the entire training set in this method, based on samples that the models have not yet seen. This method can be used for self-thought scoring as well as a transfer learning scoring method. If the model itself is used to calculate scores (ST), we called this method {\it cross validated self-thought } (CVST) scoring function. When a different model is used to calculate scores (TL), this method is referred to as a {\it cross validated transfer learning } (CVTL) scoring function.

\subsubsection{Model Based Ensemble Scoring Functions}
Ensemble learning is one of the most important validated methods to increase the success of machine learning models. Basically, it increases learning performance by using a base learner whose predictions are different from each other.

When the deep learning model starts with different initial weights, the training is completed at different points.  Combining predictions yields more successful results than predictions from a single model. Based on this concept, we considered integrating the scores generated by the deep learning models. We took the arithmetic mean of the scores generated with various initial weights. Let $R_n$ be a sequence of scores obtained from different methods. The final ensemble scores are calculated by $e_i = \frac{R_i}{\sum R_i} $ . $ {R_i} $ values can be calculated by ST, TL, CVST and CVTL methods. The ensemble versions of these methods are named as follows. {\it Ensemble self-thought } (EST), {\it ensemble transfer learning } (ETL), {\it ensemble cross validated self-thought } (ECVST) and {\it ensemble cross validated transfer learning } (ECVTL) scoring functions. To generate the EST and ETL scores, both ST and TL are trained 5 times with different initial weights. To create cross-valitated ensemble scores (ECVST, ECVTL), models are trained 5 times both with different initial weights and different splits.

\subsection{Text Based Scoring Functions}
For natural language processing tasks; Instead of model-based score functions, text-based scores can be used. Scores can be determined according to the characteristics of the sentences. We used 2 different methods to construct scoring functions with sentence characteristics. These methods are sentence length and n-gram based properties. 

\subsubsection{Sentence Length Scoring Function}

One can argue that longer sentences are more difficult because it usually consists of small parts that need to be understood. However, it can be said that long sentences give more clues about the meaning of the sentence, and it can be said that long sentences are easier. For this reason, we tested both arguments in the experiments. 
Let $s_i$ sequences of words $s_i = \{ w^i_0,...,w^i (N_i)  \}$. If the long sentences are assumed as easy samples, sentence scores are calculated as $k_i = length(s_i)$ and $r_i = \frac{k_i}{\sum k_i} $. For using short sentences first, ${k_i}$ is reversed as $1 / k_i$ before normalized. 

\subsubsection{N-gram Based Scoring Functions}
This method calculates sentence difficulty for each sentence using unigram, bigram and trigram probabilities \cite{campos2021curriculum}. Based on the intuition that the model will struggle to understand the word unless it has seen it in the corpus or there is insufficient diversity to infer its meaning. Presenting to the model to rare words early may result in poorly estimated word representations, as word embeddings can have high variance when rare words are recognized early in the model. Calculation of unigram, bigram, trigram probabilities given in Equation 1 where $uc, bc, tc$ are the total counts of unigram, bigram and trigram in the corpus. C is corpus, $c(y)$ represents count of $y$, $x \in C$  sample in the corpus. $w_i \in x$ is a word in a sentence, and $l(.)$ is the length of a sentence in n-grams.

\begin{equation}
\begin{gathered}
p(w_n) = \frac {\sum_{x \in C } {c(w_n)}  } {uc} \\
p(w_n,w_m) = \frac {\sum_{x \in C } {c(w_n, w_m)}  } {bc} \\
p(w_n,w_m,w_j) = \frac {\sum_{x \in C } {c(w_n,w_m,w_j)}  } {tc} \\
unigram\text{-}\epsilon (s_i) = - \sum_{n=0}^{l(s_i)} p(w_n) * \log(p(w_n)) \\
bigram\text{-}\epsilon (s_i) = - \sum_{n=0}^{l(s_i)-1} p(w_n,w_{n-1}) * \log(p(w_n,w_{n-1})) \\
trigram\text{-}\epsilon (s_i) = - \sum_{n=0}^{l(s_i)-2}  p(w_n,w_{n-1},w_{n-2}) * \log(p(w_n,w_{n-1},w_{n-2}))
\end{gathered}
\end{equation}


\subsection{Pacing Functions}
In CL, the {\it pacing function } defines how many training samples will be used in the training stages in the corresponding epoch. It is usually defined as a monotonically non-decreasing function. Since the aim of this study was to examine the different scoring functions, the same pacing function was used in the experiments.

The following is the definition of the pacing function used. Let the training set to be used consists of 3n samples and 3t epochs will be trained. The pacing function's first t values are n. The next t values are 2n. The remaining elements are 3n. In other words, after sorting the training set from easy to difficult, it is divided into three equal-sized subsets. The first subset is used for the first t epoch. Following the t epoch, the two easiest subsets are combined. Finally, the entire dataset is used for t epochs.

\subsection{CL Training Algorithms}
We define 2 types of CL training algorithms. These are Greedy Curriculum Learning (GCL) and Probabilistic Curriculum Learning (PCL). When selecting training subsets, both algorithms uses entire dataset's class distributions. It means keeping samples at the same ratio of samples from each class as in the training set.

After sample difficulties determined, GCL uses the simplest n samples. It continues to use the same instances until new data is added. The number n and how much new data is going to add are determined by the pacing function. This process continues until all training data is used. PCL on the other hand, instead of using simplest n samples, it selects n samples with probabilistically proportional to the relevant scoring function. n samples vary after each epoch. Other phases of training progress like GCL. GCL and PCL pseudocodes are given in Algorithm-1 and Algorithm-2. 

\begin{algorithm}
\caption{GCL Training}
\begin{algorithmic}[1]
\Require
\State $M$ : Model to train.
\State $D$ : Training dataset, consists of X: Input and y: Output.
\State $f$ : Scoring function.
\State $g$ : Pacing function.
\State $T$ : Epoch count for training.
\Statex

\Procedure {train}{}
\State Sort $D$ according to $f$, in descending order
\For {$i \in \{1...T\}$}
\State k = $g_t(i)$
\State $D'_i$ = $D[1...k]$
\State \Call{train\_model}{$M$,$D'_i$} (Perform one epoch optimization for $D'_i$)
\EndFor

\EndProcedure
\end{algorithmic}
\end{algorithm}

\begin{algorithm}
\caption{PCL Training}
\begin{algorithmic}[1]
\Require
\State $M$ : Model to train.
\State $D$ : Training dataset consists of X: Input and y: Output.
\State $f$ : Scoring function.
\State $g$ : Pacing function.
\State $T$ : Epoch count for training.
\Statex

\Procedure {train}{}
\For {$i \in \{1...T\}$}
\State k = $g_t(i)$
\State $n$ $\xleftarrow{}$ randomly pick k samples indices from distribution f
\State \Call{train\_model}{$M$,$D(n)$} (Perform one epoch optimization for $D(n)$)
\EndFor

\EndProcedure
\end{algorithmic}
\end{algorithm}

\section{Experiments}

Experiments are divided into two parts as Image Classification and Text Classification.

\subsection{Image Classification}
To test scoring functions and CL algorithms, we selected 4 image classification datasets. Datasets are cifar10 \cite{krizhevsky2009learning}, cifar100 \cite{krizhevsky2009learning}, fashion mnist (fmnist) \cite{xiao2017fashion}, kmnist \cite{clanuwat2018deep}. While cifar10 and cifar100 contains 50000 training samples, fmnist and kmnist contain 60000 training samples. Each dataset includes a total of 10000 test samples. While the cifar100 dataset has 100 classes, the other datasets have 10 classes. 

We used a basic CNN model to train \cite{tensorflow}. We used batch normalization before each activation layer in the model. For transfer scoring function we used \cite{ektasharma_2020} model. Whereas the transfer learning model has a total of 552,874 parameters to train, the self-thought model has 123,466 parameters to train. During training model we used Adam optimizer for all datasets. Also, batch size 128 was used for all data sets. To determine the number of epochs required to train the models, we performed the following steps. We continued training until there was no improvement in the validation dataset. We then trained the models with 3 times that number of epochs. We used the same epoch number to train both the vanilla and CL techniques. We see that the potential maximum success of models with long-term training.

We determined the success criteria as the top-1 accuracy measure in the test set. For this, we tested the models at the end of each epoch. As a result of these tests, we determined the maximum accuracy reached as the success of the model. Since deep learning algorithms are stochastic algorithms, they lead to different results in each trial. For this reason, we trained each model with different initial weights 5 times and got their average success.

\begin{table}[H]
\centering
\caption{Average Maximum Accuracy Table \\
Table 1 shows that average accuracies over 5 trials. The above values are the rankings created with the self-thought scoring functions; below shows the results of the rankings created with the transfer scoring functions.}
\centering
\begin{tabular}{@{}lllll@{}}
\toprule
\multicolumn{1}{r}{} & \multicolumn{1}{r}{cifar10} & \multicolumn{1}{r}{cifar100} & \multicolumn{1}{r}{fmnist} & \multicolumn{1}{r}{kmnist} \\ \midrule
Vanilla              & 71.76                       & 38.61                        & 91.19                      & 96.22                      \\
Rand-CL              & 69.59                       & 32.1                         & 90.92                      & 96.2                       \\ \midrule
ST-GCL               & 72.6                        & 35.64                        & 91.21                      & 96.02                      \\
ST-PCL               & 72.32                       & 38.03                        & 91.16                      & 96.28                      \\
EST-GCL              & 72.7                        & 37.02                        & 91.25                      & 95.96                      \\
EST-PCL              & 73.09                       & 39.14                        & 91.48                      & 96.18                      \\
CVST-GCL             & 72.45                       & 35.53                        & 91.2                       & 95.91                      \\
CVST-PCL             & 72.33                       & 37.72                        & 91.25                      & 96.17                      \\
ECVST-GCL            & 72.85                       & 36.64                        & 91.46                      & 95.9                       \\
ECVST-PCL            & 73.24                       & 39.33                        & 91.49                      & 96.09                      \\ \midrule
TL-GCL               & 73.63                       & 36.64                        & 91.46                      & 96.03                      \\
TL-PCL               & 73.7                        & 39.33                        & 91.71                      & 96.39                      \\
ETL-GCL              & 73.83                       & 37.43                        & 91.49                      & 95.98                      \\
ETL-PCL              & 74.09                       & 39.41                        & 91.52                      & 96.29                      \\
CVTL-GCL             & 73.47                       & 38.06                        & 91.45                      & 96.24                      \\
CVTL-PCL             & 73.67                       & 39.77                        & 91.57                      & 96.2                       \\
ECVTL-GCL            & 74.13                       & 37.51                        & 91.41                      & 95.84                      \\
ECVTL-PCL            & 74.0                        & 39.76                        & 91.62                      & 96.18                      \\ \hline           

\end{tabular}

\end{table}

Table-1 shows the results. The meanings of the abbreviations in the table are as follows; Greedy Curriculum Learning (GCL), Probabilistic Curriculum Learning (PCL), self-thought (ST), ensemble self-thought (EST), transfer scoring (TC), ensemble transfer scoring (ETC) cross-validated self-taught (CVST), ensemble cross-validated self-taught (ECVST), cross-validated transfer learning (CVTL), ensemble cross-validated transfer learning (ECVTL). These results show that; in general, the scores in the lower half of the table are higher than the upper half. This shows that TL is more successful than ST. Using another more successful model to determine the ranking may increase success. Of course, this situation is not always applicable as it requires applying to another model. However, this could indicate that potentially even better rankings are possible. In the self-thought method, although the success rates do not change much from ST-GCL with individual methods, combining them with ECVST-PCL  gives the most successful results. We also experimented with anti-curriculum versions of these methods. But anti-curriculum versions performed significantly worse than both the curriculum and vanilla method. In three of the four datasets, the ECVST-PCL results are statistically significant compared to vanilla and ST-GCL. In addition, the PCL algorithm gave more successful results than the GCL algorithm.

\subsection{Text Classification}
We collected some dataset model pairs from various sources on the Internet for the text classification task. We used the 20-news\cite{20_news,20_news_model} and reuters \cite{reuters,reuters_model} datasets-model pairs for news classification tasks. The 20-news dataset mainly uses pre-trained glove embeddings in training. Reuters dataset on the other hand, uses dense layers to classify. 20-news dataset contains 20 class and Reuters dataset contains 46 classes. Sarcasm detection is another task. The purpose of this dataset is to determine whether news headlines contain sarcasm \cite{sarcasm,book}. It's a binary classification problem. The main layer used in the model \cite{sarcasm_model} is bidirectional-gru. Last dataset-model pair we picked is hotel reviews sentiment analysis. In this dataset includes reviews and scores about hotels \cite{hotel}. The purpose of the created model \cite{hotel_model} is to determine the score of the reviews. It consists of 5 classes in total. 

In NLP datasets, we did not use transfer scoring functions because the models were large enough. We use both model-based self-thought and text-based sorting. In addition to self-thought, the following scoring functions are used: sentence length, unigram, bigram, trigram scores, and ensemble scores.

\begin{figure}[h]
\includegraphics[width=\columnwidth]{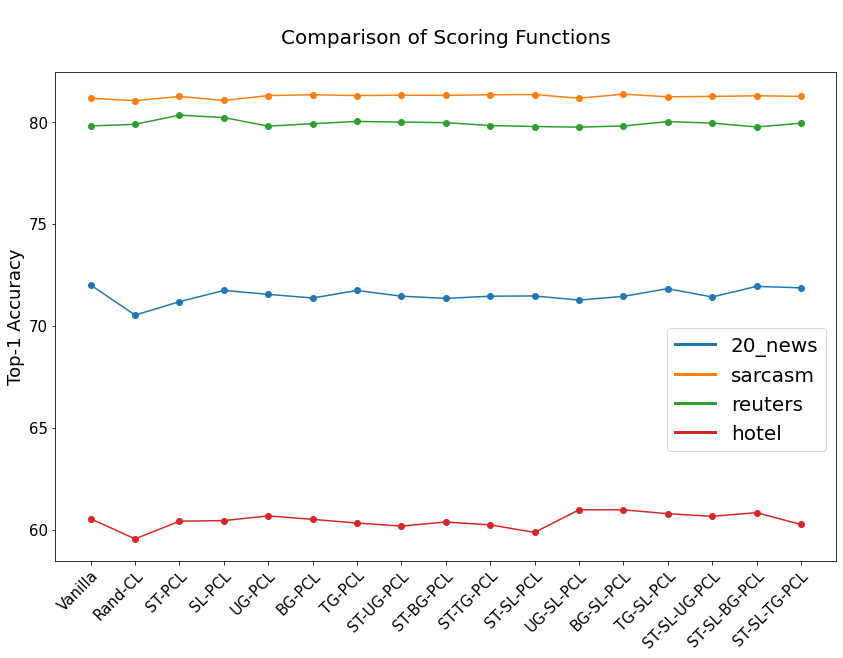}
\caption{Comparison of Scoring Functions on NLP Datasets}
\label{fig:comp}
\end{figure}

In Figure -1, comparisons of the methods are given in 4 datasets.
Abbreviations in the figure are as follows; Self-Thought (ST), Sentence Length (ST), unigram (UG), bigram (BG), and trigram (TG). For ensemble versions, abbreviations are used together in the methods used in the ensemble.

All scoring functions were trained five times with GCL and PCL. In these experiments, PCL performed consistently better than GCL, so we included only PCL in Figure 1. In text classification experiments, a method that performs well against vanilla from text-based and model-based scoring functions was not found.

\section{Discussion and Conclusions}
The purpose of this research is to analyze various scoring functions in Curriculum Learning. Different scoring functions and their ensemble versions are used. Experiments were conducted for image and text classification tasks. Three of the four datasets CL showed significant performance gains compared to vanilla training, which did not include any sorting or subsets in the image classification datasets. The highest performance improvements were obtained with the transfer learning scoring functions, obtained from a more successful model than the model to be trained. It provides significant improvements in methods that do not use transfer learning for sequencing. For ranking, the ensemble scoring functions using the model itself (self-thought) performed best among the methods that did not rely on any external source.  Transfer learning scoring functions reveal the potential for better ranking. Model-based and text-based and ensemble version scoring functions are used in text classification datasets. In contrast to image classification datasets, there was no consistent improvement with a single scoring function. It shows that different new scoring functions are waiting to be found for text classification tasks.

\section*{Acknowledgment}
This study was supported by the Scientific and Technological Research Council of Turkey (TUBITAK) Grant No: 120E100.

\bibliography{citation}
\bibliographystyle{IEEEtran}

\end{document}